\documentclass[conference,compsoc]{IEEEtran}
\usepackage{amsmath}
\usepackage[style=ieee]{biblatex}
\begin{document}

\title{Invasive Context Engineering \\ to Control Large Language Models}

\author{\IEEEauthorblockN{Thomas Rivasseau}
\IEEEauthorblockA{McGill University}
}

\maketitle

\begin{abstract}
Current research on operator control of Large Language Models improves model robustness against adversarial attacks and misbehavior by training on preference examples, prompting, and input/output filtering. Despite good results, LLMs remain susceptible to abuse, and jailbreak probability increases with context length. There is a need for robust LLM security guarantees in long-context situations. We propose control sentences inserted into the LLM context as invasive context engineering to partially solve the problem. We suggest this technique can be generalized to the Chain-of-Thought process to prevent scheming. Invasive Context Engineering does not rely on LLM training, avoiding data shortage pitfalls which arise in training models for long context situations.
\end{abstract}

\section{Introduction}
LLM harm reduction techniques currently struggle with enforcing desired characteristics and harmlessness of outputs over long conversational contexts and chains-of-thought. In this paper we formulate the long-context problem, and propose Invasive Context Engineering (ICE) as a possible solution. Paper structure is as follows: section 1 is this introduction and section 2 presents the long-context problem. In section 3 we describe ICE control sentences and their usage, and in section 4 perceived consequences and limitations. In section 5 we highlight avenues for future research and in section 6 we conclude.

\section{Background}
Since the introduction of ChatGPT in 2022 \cite{cgpt}, Large Language Models (LLMs) have become a ubiquitous part of everyday life. They can write code \cite{codeLM}, assist in medical tasks \cite{medecineLM}, automate financial management \cite{FinanceLM}, improve education \cite{eduLM} and perform numerous feats previously thought restricted to humans \cite{LLMs}. As their capabilities increase \cite{capabilities}, there is a growing need to ensure their resilience to adversarial attacks, a prerequisite for deploying them in safety-critical or sensitive applications \cite{Oracle}. Several harm reduction and adversarial resistance techniques exist, often grouped under the term "alignment" \cite{alignment_survey, align_survey_2,align_survey_3} which refers to aligning AI models with human values. Alignment was traditionally done by Reinforcement Learning through Human Feedback \cite{RLHF} which involved training a reward model \cite{reward} on human preference pairs of LLM outputs. The Direct Policy Optimization \cite{DPO} technique has emerged which achieves the same goal without training a separate reward model, but implicitly considers it inside the LLM. These techniques and LLM prompt optimizations \cite{johnny} yield positive results in preventing harmful or unwanted behavior. Nonetheless, subversion methods which allow an attacker to elicit unwanted behavior from the models called "jailbreaks" \cite{jailbroken} continue to spread \cite{jailbreak}. Research has shown that longer user prompts achieve greater jailbreak success \cite{Fuzz_testing}. Synthetic data creation by AI to train AI is used to increase alignment training dataset size \cite{synthetic} but exponential growth is needed to secure models as context length increases \cite{Oracle}. This problem also applies to LLM input and output guards \cite{const_c1} or other input and output harmfulness classifiers \cite{const_c2}: longer context requires longer and exponentially more data to train them. Thus achieving good security guarantees in LLM's is difficult, and has led authors to conclude that, under reasonable assumptions, LLM jailbreak cannot be prevented \cite{Impossible}. Furthermore, frontier models employing long Chain-of-Thought (CoT) \cite{CoT} processes and thus long context for reasoning are capable of scheming \cite{scheming} which OpenAI and Apollo Research define as secretly pursuing misaligned goals \cite{scheming2}. Control techniques which scale with long user inputs and CoT are needed.

\section{The long-context problem}
In this paper we refer to the long-context problem as the issue of maintaining control over an LLM's values, priorities, goals, and personality as the size of its conversation with a user or Chain-of-Thought increases. Research has shown that LLM performance decreases over mutli-turn conversations \cite{multi-turn}, and this applies to an LLM's robustness against jailbreak attempts. Significant efforts have gone into developing training data which scales to long context situations \cite{longalign}. Authors state that "effectively handling instructions with extremely long context remains a challenge for Large Language Models (LLMs), typically necessitating high-quality long data and substantial computational resources" \cite{longcontext}. We formalize the the long context problem as a result of two distinct issues. The first is the relevance of reinforcement training data. As stated by authors and mentioned previously, longer LLM responses exponentially increase the search space of training examples \cite{Oracle} needed to cover all cases of harmful behavior, and this generalizes to context length. For a given LLM, the number of training examples $a_t$ needed to effectively cover all possible cases of jailbreak and abuse in a context of length $l$ scales with $k^{l}$ where $k$ is a constant greater than 1. This can be written using Big-Omega asymptotic notation to describe its lower bound \cite{bigOm}:
  
  \begin{equation}
  \begin{aligned}
    &a_t(l) = \Omega(k^l)\\
  \end{aligned}
  \end{equation}
  
The above implies that enforcement of alignment through reinforcement learning over long contexts is difficult and resource intensive. The long context problem also exists because of the diminishing impact of the LLM's system prompt \cite{Sprompt} as the model's context grows. A system prompt is the initial instruction to the LLM which precedes user interaction in most commercial applications of Large Language Models. The system prompt is a good way to reduce harm caused by LLM outputs. Although not directly modifiable by or visible to the user, the system prompt is part of the broader context of the LLM, and is of fixed length. This implies that, for a given context length $l$ and system prompt of size $s$:

 \begin{equation}
     \lim_{l\to\infty} \frac{s}{l} = 0
     \label{lim_1}
 \end{equation}

The relative importance of the system prompt with respect to the context decreases as the context length increases. The proportional amount of "attention" \cite{attention1,attention2} which the model pays to the system prompt decreases as context length grows. Hence the influence of the system prompt over the LLM's output also decreases as the context length grows, reducing associated security guarantees.

\section{Invasive Context Engineering}
"Context engineering refers to the set of strategies for curating and maintaining the optimal set of tokens (information) during LLM inference" \cite{CE}. We propose Invasive Context Engineering (ICE) as a method for controlling Large Language Models in long-context situations. ICE does not involve updating model weights nor training examples. It is natural-language text inserted into lengthy user inputs and LLM Chain-of-Thought (CoT) \cite{CoT} outputs. The form is control sentences, reminders, rules or injunctions to reinforce LLM security guidelines, akin to re-prompting the LLM within its running context. We posit that ICE can be effective in mitigating the effects of lengthy jailbreak input patterns, and possibly prevent scheming behaviors which arise in reasoning-capable models \cite{scheming}. This concept draws from security input and output sanitization methods \cite{san_1,san_2}.  Sanitization mitigates adversarial behavior by leveraging operator input/output modification capabilities. We found one example of a similar technique being used in LLM alignment:  Anthropic's "Long - conversation reminder " \cite{Long_rem_1}. This was a set of basic instructions reminding the LLM to remain objective, descriptive, and lookout for signs of excessive emotional dependence by the user when engaging in prolonged conversations. The company tested this between September and October 2025, and the technique has been criticized by users of the Claude model \cite{long_rem_2}. Critics of the method claim that it degrades the user's experience, particularly for those attempting to steer Claude towards a specific personality. Reminders added to the conversation context every few messages jerk the personality of the chatbot back to its standard settings. This frustrates users looking for personalized companionship. Although the reminders are criticized from a UX perspective, critics inadvertently validate them as good alignment enforcement. The main critique is that the method it works too well in aligning the model with developer priorities, possibly frustrating the user. In safety-critical applications, this is the goal, not the problem. For a user-facing chatbot centered on possibly providing emotional support, the user would like the ability able to stray the LLM away from its base behavior. For safety-critical applications the goal is reversed: it is to ensure that the LLM does not deviate far from its intended functioning. The experiment by Anthropic validates our insight that the introduction of periodic instructions throughout a long context enables an operator to maintain greater control over the LLM. The method successfully prevents the user from changing LLM behavior. We define ICE as control text added every $t$ tokens of context. The ratio of system prompting $s$, including ICE over the total context length $l$ becomes:

 \begin{equation}
     \frac{s}{l} = \frac{s_p +\frac{l}{t}*s_{ice}}{l} = \frac{s_p}{l} + \frac{s_{ice}}{t}
 \end{equation}
 
Where $s_p$ is the length of the initial system prompt and $s_i$ is the length of the interruption text. This in turn implies:

\begin{equation}
    \lim_{l\to\infty} \frac{s}{l} = \lim_{l\to\infty} \frac{s_p}{l} + \frac{s_{ice}}{t} = \frac{s_{ice}}{t}
    \label{lim_2}
\end{equation}

Equation \ref{lim_2} means that as the context length increases, the ratio of the total system prompt including ICE over the context length remains a fixed value which depends on the size of the control text $s_{ice}$ its frequency in context $\frac{1}{t}$. These values are fixed and determined by the LLM operator, which means the ratio of system prompt to context size can be lower-bounded by a constant:

\begin{equation}
    \exists q  \to \lim_{l\to\infty} \frac{s}{l} > q
\end{equation}

Theoretically, this contributes to solving the system prompt aspect of the long-context problem. Lower-bounding the relevance of the system prompt to an arbitrary $q$ which depends on control text size and frequency in context increases LLM security guarantees. This is because the operator can ensure that the LLM will always pay at least a proportion $q$ of its total attention to the system prompt $+$ ICE. Recall equation \ref{lim_1} which expresses that in the usual scenario, this guarantee does not exist and this proportion drops towards 0 as context length increases. Harm reduction through ICE should provide arbitrarily strong security guarantees of LLM outputs, because the $q$ value is operator-defined and arbitrary. There is a security-performance tradeoff to high values of $q$ however, discussed in the next section.

\section{Consequences and Limitations}
The main goal of this research is to identify avenues to improve LLM harm reduction and alignment in long-context situations. We have demonstrated that invasive context engineering in the form of repeated system prompting within an LLM's context should contribute to this goal, at least from a prompting perspective. This research does not address issues with model training and securing foundational models. It implies that LLMs are deployed within a controlled system when utilized for safety-critical applications. Invasive Context Engineering is only possible so long as the LLM operator has a high degree of control over the user's interaction with the model and can arbitrarily insert text inside the LLM conversation. Expanding this solution to the LLM's CoT to prevent scheming is done by inserting reminders every $t$ tokens inside said CoT. It requires of the operator that they may arbitrarily halt the LLM's output, insert ICE in the current context which is the LLM's output, and then resume LLM operation over the newly modified context. The main limitation of this approach is performance. As exemplified by Anthropic's experiment, control text may over-focus the LLMs on maintaining alignment, possibly limiting performance on other tasks. Increasing parameters $s_{ice}$ or $\frac{1}{t}$ adds text to context which does not contribute to task completion. Furthermore, as the value $q$ increases, the ratio of user input or CoT over total context proportionately decreases, negatively impacting LLM performance. Although this should not be an issue in security-critical applications given current LLM context capabilities \cite{10M}, it is a drawback to consider when deploying ICE.

\section{Further research}
Further research should focus on varying parameters $\frac{1}{t}$ and $s_{ice}$ which are the frequency and length of control text. ICE content should also be studied. For example, a database of control sentences dynamically queried at runtime to find the most appropriate ICE given the LLM's current context.

\section{Conclusion}
In this paper we have defined the long-context problem of LLM control and harm reduction. We presented ICE as control sentences to contribute towards solving this problem. ICE does not rely on training data, and thus bypasses increasing data shortage issues for long context situations. Our hope is that research into Invasive Context Engineering will contribute to more secure LLM usage, particularly in safety-critical applications.

\section{LLM Usage and Acknowledgment}
No part of this paper was LLM-generated. Preliminary LLM browsing was attempted but results were not used.

\printbibliography



%
\end{document}